\documentclass[11pt,a4paper]{article}
\usepackage[hyperref]{acl2020}
\usepackage{times}
\usepackage{latexsym}
\usepackage{acro}
\usepackage{makecell}
\usepackage{graphicx}
\usepackage{multirow}
\usepackage{xcolor}\usepackage{xcolor}

\newcommand\model[1]{#1}
\usepackage{todonotes}
\usepackage{amsmath}
\usepackage{subcaption}
\usepackage{footmisc}

\usepackage{relsize,etoolbox}
\AtBeginEnvironment{quote}{\smaller}

\usepackage{microtype}

\aclfinalcopy 

\DeclareAcronym{sd}{
	short = StD,
	long = Stance Detection
}

\title{Stance Detection Benchmark: How Robust Is Your Stance Detection?}

\author{Benjamin Schiller and Johannes Daxenberger and Iryna Gurevych \\
  Ubiquitous Knowledge Processing Lab (UKP-TUDA) \\
  Department of Computer Science, Technische Universit{\"a}t Darmstadt \\
  \url{www.ukp.tu-darmstadt.de} }

\date{}

\begin{document}
\maketitle
\begin{abstract}
%
%
\ac{sd} aims to detect an author's stance towards a certain topic or claim and 
has become a key component in applications like fake news detection, claim validation, and argument search.
However, while stance is easily detected by humans, machine learning models are clearly falling short of this task.
Given the major differences in dataset sizes and framing of \ac{sd} (e.g. number of classes and inputs), 
we introduce a StD benchmark that 
learns from ten \ac{sd} datasets of various domains in a \textit{multi-dataset learning} (MDL) setting, as well as from related tasks via transfer learning.
Within this benchmark setup, we are able to present new state-of-the-art results on five of the datasets.
Yet, the models still perform well below human capabilities and even simple adversarial attacks severely hurt the performance of MDL models.
Deeper investigation into this phenomenon suggests the existence of biases inherited from multiple datasets by design.
Our analysis emphasizes the need of focus on robustness and de-biasing strategies in multi-task learning approaches.
The benchmark dataset and code is made available.\footnote{\url{https://github.com/UKPLab/mdl-stance-robustness}\label{fn_repo}}
\end{abstract}

\section{Introduction}

\acf*{sd} represents a well-established task in natural language processing and is often described by having two inputs; (1) a topic of a discussion and (2) a comment made by an author.
Given these two inputs, the aim is to find out whether the author is in favor or against the topic.
For instance, in SemEval-2016 Task 6 \citep{mohammad2016semeval}, the second input is a short tweet and the goal is to detect, whether the author has made a positive or negative comment towards a given controversial topic:

\begin{quote}
\textbf{Topic}: \textit{Climate Change is a Real Concern} \\
\textbf{Tweet}:  \textit{Gone are the days where we would get temperatures of Min -2 and Max 5 in Cape Town \#SemST}\\
\textbf{Stance}:  \textit{FAVOR}
\end{quote}
The task has a long tradition in the domain of political and ideological online debates \citep{mohammad2016semeval,walker2012evidence,somasundaran2010recognizing,thomas2006get}. In recent years, it has been brought into the focus of attention by the uprising debates around fake news, where \ac{sd} is an important pre-processing step \citep{fnc2017,derczynski2017semeval,ferreira2016emergent}, as well as for other downstream tasks like argument search \citep{stab-etal-2018-cross} and claim validation \citep{popat2017truth}.
As such, high performance in \ac{sd} is a crucial step in successfully leveraging machine learning (ML) for argumentative information retrieval and fake news detection. 

However, while humans are quite capable of assessing correct stances, ML models are often falling short of this task (see Table \ref{table_agreement}).
\begin{table}[!b]
\centering
\resizebox{0.47\textwidth}{!}{
\def\arraystretch{1.3}
\begin{tabular}{lcc}
\Xhline{2\arrayrulewidth}
\textbf{Dataset} & \textbf{State-of-the-art} & \textbf{Agreement}  \\\hline
\makecell[l]{ARC* \citep{habernal2018argument}} & 57.30\% & 77.30\% \\
FNC-1 \citep{fnc2017} & 61.10\% & 75.40\% \\
PERSPECTRUM \citep{chen2019perspectrum} & 70.80\% & 90.90\%  \\
\Xhline{2\arrayrulewidth}
\end{tabular}
}
\caption{Inter-annotator agreement (IAA) vs. state-of-the-art results. ARC/FNC-1 in F$_1$ macro, PERSPECTRUM in F$_1$ micro. *IAA in \citet{hanselowski2018fnc}.}
\label{table_agreement}
\end{table}
As there are numerous domains to which \ac{sd} can be applied, definitions of this task vary considerably. For instance, the first input can be a short topic, a claim, or sometimes is not given at all, while the second input can be another claim, an evidence, or even a full argument.
Further, the second input can differ in length between a sentence, a short paragraph, and whole documents.
The number of classes can also vary between 2-class problems (e.g. \textit{for}/\textit{against}) and more fine-grained 4-class problems (e.g. \textit{comment}/\textit{support}/\textit{query}/\textit{deny}). Moreover, the number of samples varies drasticially between datasets (for our setup: from 2,394 to 75,385).
While these differences are problematic for cross-domain performance, it can also be seen as an advantage, as it concludes in an abundance of datasets from different domains that can be integrated into transfer or multi-task learning approaches.
Yet, given the decent human performance on this task, it is hard to grasp why ML models fall short of \ac{sd}, while they are almost on par at related tasks like Sentiment Analysis\footnote{\url{http://nlpprogress.com/english/sentiment_analysis.html}} and Natural Language Inference\footnote{\url{http://nlpprogress.com/english/natural_language_inference.html}} (NLI).

Within this work, we provide foundations for answering this question.
We empirically assess whether the abundance of differently framed \ac{sd} datasets from multiple domains can be leveraged by looking at them in a holistic way, i.e. training and evaluating them collectively in a multi-task fashion. However, as we only have one task but multiple datasets, we henceforth define it as multi-dataset learning (MDL).
And indeed, our model profits significantly from  datasets of the same task via MDL with +4 percentage points (pp) on average, as well as from related tasks via transfer learning (TL) with +3.4pp on average.

However, while we gain significant performance improvements for \ac{sd} by using TL and MDL, the expected robustness of these approaches is missing.
We show this using a modified version of the \textit{Resilience} score by \citet{thorne-etal-2019-evaluating} which reveals that TL and MDL models are even less robust than single-dataset learning (SDL) models.
We investigate this phenomenon through low resource experiments and observe that less training data leads to an improved robustness for the MDL models, narrowing down the gap to the SDL models.
We thus assume that lower robustness stems from dataset biases introduced by the vast amount of available training data for the MDL models, leading to overfitting.
Consequently, adversarial attacks that target such biases have a more severe impact on models that had more biased training data and overfitted on these biases.


The contributions of this paper are as follows:
(1)~To the best of our knowledge, we are the first to combine learning from related tasks (via TL) and MDL, designed to capture all facets of \ac{sd} tasks, and achieve new state-of-the-art results on five of ten datasets. 
(2)~In an in-depth analysis with adversarial attacks, we show that TL and MDL for \ac{sd} generally improves the performance of ML models, but also drastically reduces their robustness if compared to SDL models. 
(3)~To foster the analysis of this task, we publish the full benchmark system including model training and evaluation, as well as the means to add and evaluate adversarial attack sets and low resource experiments.\footref{fn_repo}
All datasets, the fine-tuned models, and the machine translation models can be automatically downloaded and preprocessed for consistent future usage.

\begin{table*}[!hbt]
\centering
\resizebox{0.99\textwidth}{!}{
\def\arraystretch{1.3}
\begin{tabular}{lllll}
\Xhline{2\arrayrulewidth}
\textbf{Dataset groups} & \textbf{Domain}  & \multicolumn{3}{l}{\textbf{Example}} \\ \hline
&& \textbf{Topic} & \textbf{Comment} & \textbf{Stance} \\\hline
\makecell[l]{ibmcs} & Encyclopedia & [...] atheism is the only way&Atheism is a superior basis for ethics & PRO\\\hline
\makecell[l]{semeval2019t7\\semeval2016t6} & Social media & \makecell[l]{(Charlie Hebdo),\\Feminist Movement}&\makecell[l]{"[...] suspected \#CharlieHebdo gunmen have been killed" yayyy\#boom\\I believe that every women should have their own rights!! \#SemST} & \makecell[l]{Support\\Favor} \\\hline
\makecell[l]{fnc1\\snopes} & News &\makecell[l]{Russia can be a partner\\Farmers feed their cattle candy [...]
}&\makecell[l]{If America, Russia and Iran can come together [...] it will be a start\\The alternative would be to put [the candy] in a landfill somewhere.}& \makecell[l]{Agree,\\Agree}\\\hline
\makecell[l]{scd\\perspectrum\\iac1\\arc} & \makecell[l]{Debating\\forums} & \makecell[l]{(Obama)\\School Day Should Be Extended\\existence of god\\Salt should have a place at the table}&\makecell[l]{I think Obama has been a great President. [...] 
\\So much easier for parents!\\ {[}...] the Bible tells me that Jesus existed, and is the Son of God [...]\\ {[}...] the iodine in salt is necessary to prevent goiter. [...]}&\makecell[l]{For\\Support\\Pro\\Agree}\\\hline
\makecell[l]{argmin} & Web search & school uniforms & We believe in freedom of choice. & CON\\\Xhline{2\arrayrulewidth}
\end{tabular}
}
\caption{All datasets, grouped by domain and with examples. Topics in parentheses signal implicit information.
}
\label{table_dataset_examples}
\end{table*}

\section{Related Work}
\textbf{Stance Detection} is a well-established task in natural language processing. Initial work focused on parliamentary debates \cite{thomas2006get} and debating portals \cite{somasundaran2010recognizing}, whereas latest work has shifted to the domain of Social Media, where several shared tasks have been introduced \cite{gorrell2019semeval,derczynski2017semeval,mohammad2016semeval}. With the shift in domains, the definition of the task also shifted: more classes were added (e.g. \textit{query} \cite{gorrell2019semeval} or \textit{unrelated}  \cite{fnc2017}), the number of inputs has changed (e.g. multiple topics for each sample \cite{sobhani-etal-2017-dataset}
),
or the definition of the inputs itself (e.g. from parliamentary speeches and debate portal posts to tweets \cite{gorrell2019semeval}, news articles \cite{fnc2017}, or argument components \cite{stab-etal-2018-cross,barhaim2017stance}). In past years, the problem of \ac{sd} has become a cornerstone for many downstream tasks like fake news detection \cite{fnc2017}, claim validation \cite{popat2017truth}, and argument search \cite{stab-etal-2018-cross}. Yet, recent work mainly focuses on individual datasets and domains. We, in contrast, concentrate on a higher level of abstraction by aggregating datasets of different domains and definitions to analyze them in a holistic way.
To do so, we leverage the idea of TL and multi-task learning (in form of MDL), as they have not only shown increases in performance and robustness \cite{ruder2017overview,weiss2016survey}, but also significant support in low resource scenarios \cite{schulz18multi}. Latest frameworks for multi-task learning include the one by \citeauthor{liu2019multi} \shortcite{liu2019multi}, which scored a new state-of-the-art on the GLUE Benchmark \cite{wang-etal-2018-glue}. In contrast to their work, we will use the framework for MDL, i.e. combining only datasets of the same task to analyze whether \ac{sd} datasets can benefit from each other by transferring knowledge about their domains.
Furthermore, we probe the robustness of the learned models to analyze whether performance increases gained through TL and MDL are in accordance with increased robustness for \ac{sd}.

\noindent  \textbf{Adversarial attacks} describe test sets aimed to discover possible weak points of ML models. While much recent work in adversarial attacks aims to break NLI systems and is especially adapted to this problem \cite{glockner2018breaking,minervini2018adversarially}, these stress tests have been applied to a wide range of tasks from Question-Answering \cite{wang2018robust} to Natural Machine Translation \cite{belinkov2017synthetic} and Fact Checking \cite{thorne-etal-2019-evaluating}. Unfortunately, preserving the semantics of a sentence while automatically generating these adversarial attacks is difficult, which is why some works have defined small stress tests manually \cite{isabelle2017challenge,mahler2017breaking}. As this is time (and money) consuming, other work has defined heuristics with controllable outcome to modify existing datasets and to preserve the semantics of the data \cite{naik-etal-2018-stress}.
In contrast to previous work, we use and analyze some of these attacks for the task of \ac{sd} to probe the robustness of our SDL and MDL models.

\section{Stance Detection Benchmark: Setup and Experiments}\label{sec:experiments}
We describe the dataset and models we use for the benchmark, the experimental setting, and the results of our experiments. For all experiments, we use and adapt the framework\footnote{\url{https://github.com/namisan/mt-dnn}} provided by \citeauthor{liu2019multi} \shortcite{liu2019multi}.

\subsection{Datasets}\label{sec:Datasets}
We choose ten \ac{sd} datasets from five different domains to represent a rich environment of different facets of \ac{sd}. Datasets within one domain may still vary by their number of classes and sample sizes. All datasets are shown with an example and their domain in Table \ref{table_dataset_examples}. In addition, Table \ref{table_datasets_stats} displays the split sizes and the class distributions of each dataset. All code to preprocess and split the datasets is available online.\footref{fn_repo} In the following, all datasets are introduced.

\noindent \textbf{arc} We take the version of the Argument Reasoning Corpus \cite{habernal2018argument} that was modified for \ac{sd} by \citeauthor{hanselowski2018fnc} \shortcite{hanselowski2018fnc}. A sample consists of a claim crafted by a crowdworker and a  user post from a debating forum. 

\noindent \textbf{argmin} The UKP Sentential Argument Mining Corpus \cite{stab-etal-2018-cross} originally contains topic-sentence pairs labelled with \textit{argument\_for}, \textit{argument\_against}, and \textit{no\_argument}. We remove all non-arguments and simplify the original split: we train on the data of five topics, develop on the data of one topic, and test on the data of two topics.

\noindent \textbf{fnc1} The Fake News Challenge dataset \cite{fnc2017} contains headline-article pairs from news websites. We take the original data without modifying it.

\noindent \textbf{iac1} The Internet Argument Corpus V1 \cite{walker2012corpus} contains topic-post pairs from political debates on internet forums. We generate a new split without intersection of topics between train, development, and test set.

\noindent \textbf{ibmcs} The IBM Debater\textregistered~- Claim Stance Dataset \cite{barhaim2017stance} contains topic-claim pairs. The topics are gathered from a debating database, the claims were manually collected from Wikipedia articles. We take the pre-defined train and test split and split an additional 10\% off the train set for development.

\noindent \textbf{perspectrum} The PERSPECTRUM dataset \cite{chen2019perspectrum} contains pairs of claims and related perspectives, which were gathered from debating websites. We only take the data they defined for the \ac{sd} task in their work and keep the exact split.

\noindent \textbf{scd} The Stance Classification Dataset \cite{hasan2013stance} contains posts about four topics from an online debate forum with all posts being self-labelled by the post's author.
The topics are not part of the actual dataset and have to be inferred from explicit or implicit mentions within a post. We generate a new data split by using the data of two topics for training, the data of one topic for development, and the data of the leftover topic for testing.

\noindent \textbf{semeval2016t6} The SemEval-2016 Task 6 dataset \cite{mohammad2016semeval} contains topic-tweet pairs, where topics are controversial subjects like politicians, Feminism, or Atheism. We adopt the same split as used in the challenge, but add some of the training data to the development split, as it originally only contained 100 samples.

\noindent \textbf{semeval2019t7} The SemEval-2019 Task 7 \cite{gorrell2019semeval}
contains rumours from reddit posts and tweets towards a variety of incidents like the Ferguson Unrest or the Germanwings crash. Similar to the scd dataset, the topics are not part of the actual dataset.

\noindent \textbf{snopes} The Snopes corpus \cite{hanselowski2019snopes} contains data from a fact-checking website\footnote{\url{www.snopes.com}} documenting (amongst others) rumours, evidence texts gathered by fact-checkers, and the documents from which the evidence originates. Besides labels for automatic fact-checking of the rumours, the corpus also contains stance annotations towards the rumours for some evidence sentences. We extract these pairs and generate a new data split.
\begin{table*}[!hbt]
\centering
\def\arraystretch{1.3}
\resizebox{1.0\textwidth}{!}{
\begin{tabular}{lccccl}
\Xhline{2\arrayrulewidth}
\textbf{Datasets} & \multicolumn{4}{c}{\textbf{\# samples}} &  \\ \hline
 & \textbf{Train} & \textbf{Dev} & \textbf{Test} & \textbf{Total} & \textbf{classes}\\ \hline
 \makecell[l]{\model{arc} (\citeauthor{hanselowski2018fnc} \citeyear{hanselowski2018fnc};\\ (\citeauthor{habernal2018argument} \citeyear{habernal2018argument}} & 12,382 & 1,851 & 3,559 & 17,792  & \makecell{unrelated (75\%), disagree (10\%), agree (9\%), discuss (6\%)} \\
\model{argmin} \cite{stab-etal-2018-cross} & 6,845 & 1,568 & 2,726 & 11,139 &   argument\_against (56\%), argument\_for (44\%) \\
\model{fnc1} \cite{fnc2017} & 42,476 & 7,496 & 25,413 & 75,385 & unrelated (73\%), discuss (18\%), agree (7\%), disagree (2\%)\\
\model{iac1} \cite{walker2012corpus} & 4,227 & 454 & 924 & 5,605 & pro (56\%), anti (34\%), other (10\%) \\
\model{ibmcs} \cite{barhaim2017stance} & 935 & 104 & 1,355 & 2,394 & pro (55\%), con (45\%) \\
\model{perspectrum} \cite{chen2019perspectrum} & 6,978 & 2,071  & 2,773 & 11,822 & support (52\%), undermine (48\%)\\
\model{scd} \cite{hasan2013stance} & 3,251 & 624 & 964 & 4,839  & for (60\%), against (40\%)  \\ 
\model{semeval2016t6} \cite{mohammad2016semeval} & 2,497 & 417 & 1,249 & 4,163   & against (51\%), favor (25\%), none (24\%) \\
\model{semeval2019t7} \cite{gorrell2019semeval} & 5,217 & 1,485 & 1,827 & 8,529   & comment (72\%), support (14\%), query (7\%), deny (7\%) \\
\model{snopes} \cite{hanselowski2019snopes} & 14,416 & 1,868 & 3,154 & 19,438 &  support (74\%), refute (26\%) \\\Xhline{2\arrayrulewidth}
\end{tabular}
}
\caption{Splits, classes, and class distributions for all used datasets.}
\label{table_datasets_stats}
\end{table*}

\subsection{Models}
\begin{figure}[!t]
\centering
{\includegraphics[width=0.41\textwidth]
{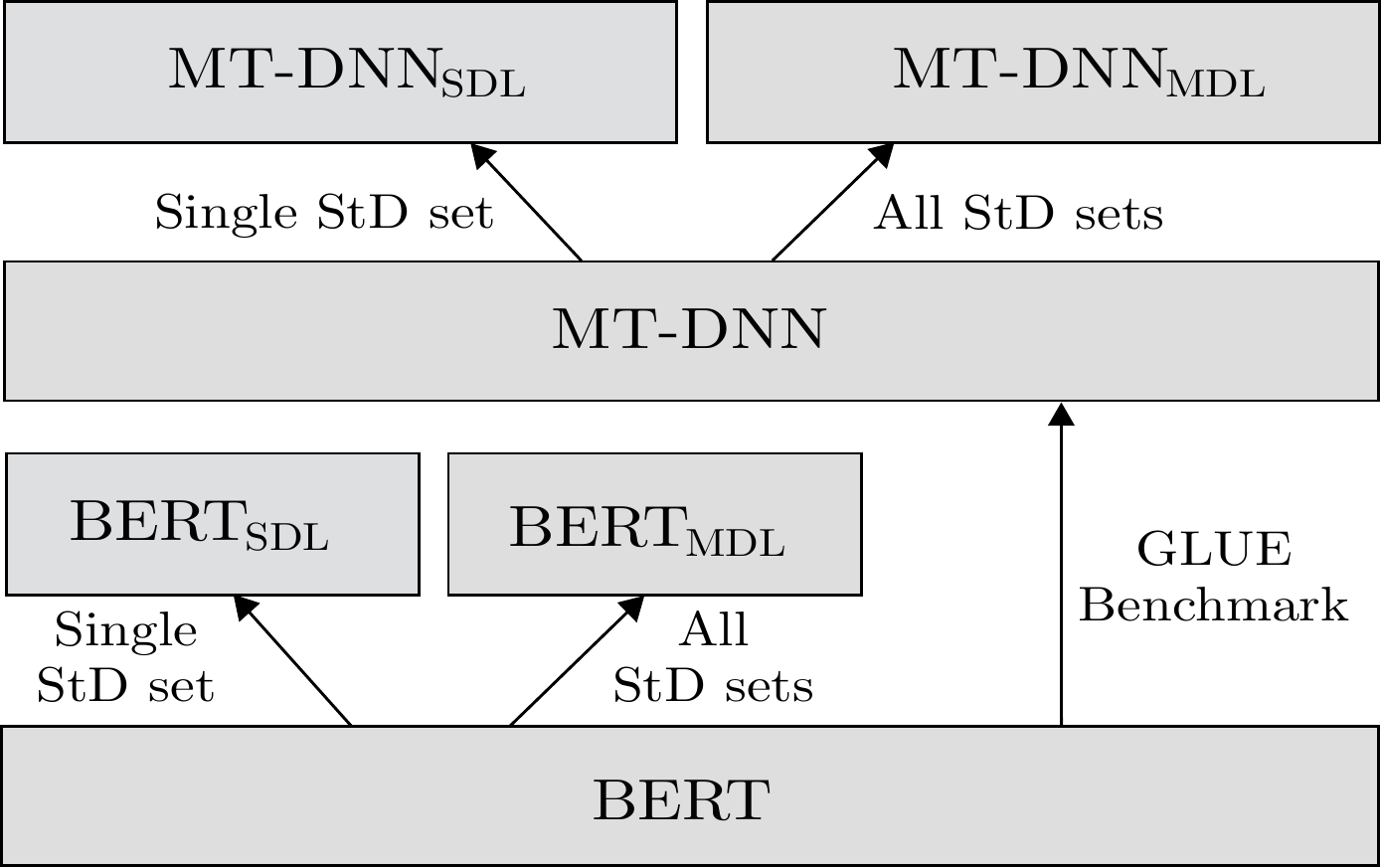}}
\caption{\label{fig_architecture} Models and their relation. Arrows symbolize training, their labels state the used training data.}
\end{figure}

We experiment on all datasets in an SDL setup, i.e. training and testing on all datasets individually, and in an MDL setup, i.e. training on all ten \ac{sd} datasets jointly. For this, we use the framework by \citeauthor{liu2019multi} \shortcite{liu2019multi}, as it provides the means to do both SDL and MDL.
The SDL is based on the BERT architecture \cite{devlin2018bert} and simply adds a dense layer on top for the classification. The MDL is also based on the BERT architecture, but each dataset has its own dataset-specific dense layer on top. While the layers of the BERT architecture are shared, the dataset-specific layers are updated for each dataset individually at training time. All datasets are batched and fed through the architecture in a random order.
As initial weights for SDL and MDL, we use either the pre-trained BERT (large, uncased) weights by \citeauthor{devlin2018bert} \shortcite{devlin2018bert} or the MT-DNN (large, uncased) weights by \citeauthor{liu2019multi} \shortcite{liu2019multi}. The latter uses the BERT weights and is fine-tuned on all datasets of the GLUE Benchmark \cite{wang-etal-2018-glue}.
By using the MT-DNN, we transfer knowledge from all datasets of the GLUE Benchmark to our models, i.e. apply TL in the form of pre-training.
Henceforth, we use SDL and MDL to define the model architecture, and BERT and MT-DNN to define the pre-trained weights of the model architecture.
This leaves us with four combinations of models: \textbf{\model{BERT$_{SDL}$}}, \textbf{\model{BERT$_{MDL}$}}, \textbf{\model{MT-DNN$_{SDL}$}}, and \textbf{\model{MT-DNN$_{MDL}$}} (see Figure \ref{fig_architecture}).

\subsection{Experimental Setting}
For all experiments in this section, we set the batch size to 16, the number of epochs to 5, and we cut each input on 100 sub-words due to hardware limitations.
Preliminary tests with the fnc1 dataset, which contains documents as one of the inputs, showed a minor drop in F$_1$ macro of less than 2pp when reducing the sequence length from 300 to 100.
To compensate for variations in the results, we train over five different fixed seeds and report the averaged results. We run all experiments on a Tesla P-100 with 16 GByte of memory. One epoch with all ten datasets takes around 1.5h.
We use the splits for training, development, and testing as shown in Table \ref{table_datasets_stats}. The table also lists the classes and class distribution for each dataset. We use the F$_1$ macro (F$_1$m$_+$) as a general metric, since the class balance for most datasets is skewed. The dataset training sizes vary from approx. 42,500 to as low as 935 samples.

\subsection{Results}
We report the results of all models and datasets in Table \ref{table_stance_results}. The last column shows the averaged F$_1$m$_+$ for a row.
We make three observations: (1) TL from related tasks improves the overall performance, (2) MDL with datasets from the same task shows an even larger positive impact, and (3) TL, followed by MDL, can further improve on the individual gains shown by (1) and (2).

We show (1) by comparing the models BERT$_{SDL}$ and MT-DNN$_{SDL}$, where a gain of 3.4pp due to TL from the GLUE datasets can be observed. While some datasets show a drop in performance, the average performance increases. We show (2) by comparing BERT$_{SDL}$ to BERT$_{MDL}$ (+4pp) and MT-DNN$_{SDL}$ to MT-DNN$_{MDL}$ (+1.8pp). The former comparison indicates that learning from similar datasets (i.e. MDL) has a higher impact than TL for \ac{sd}. The latter comparison leads to observation (3); combining TL from related tasks (+3.4pp) and MDL on the same task (+4pp), can result in considerable performance gains (+5.1pp). However, as the individual gains from TL and MDL do not add up, it also indicates an information overlap between the datasets of the GLUE benchmark and the \ac{sd} datasets.
Lastly, while BERT$_{SDL}$ already outperforms five out of six state-of-the-art results, our BERT$_{MDL}$ and MT-DNN$_{MDL}$ are able to add significant performance increases on top.

\begin{table*}[!hbt]
\centering
\resizebox{1.0\textwidth}{!}{
\def\arraystretch{1.3}
\begin{tabular}{lcccccccccc||c}
\Xhline{2\arrayrulewidth}
\textbf{Models} & \textbf{\makecell{arc\\12.4k}}  & \textbf{\makecell{argmin\\6.8k}} & \textbf{\makecell{fnc1\\42.5k}} & \textbf{\makecell{iac1\\4.2k}} & \textbf{\makecell{ibmcs\\0.9k}} & \textbf{\makecell{perspectrum\\7.0k}} &  \textbf{\makecell{scd\\3.3k}} & \textbf{\makecell{semeval2016t6\\2.5k}} & \textbf{\makecell{semeval2019t7\\5.2k}} &\textbf{\makecell{snopes\\14.4k}} & \textbf{Avg.}  \\ \hline
\textbf{\makecell[l]{Metrics\\(original)}} & F$_1$m$_+$ & F$_1$m$_+$ & \makecell{F$_1$m$_+$\\ (FNC1)} & F$_1$m$_+$& \makecell{F$_1$m$_+$\\ (Acc)} & \makecell{F$_1$m$_+$\\(F$_1$m$_-$)} & F$_1$m$_+$ & \makecell{F$_1$m$_+$\\(F$_1$m$_+\setminus$ none)} & F$_1$m$_+$ & F$_1$m$_+$ & F$_1$m$_+$ \\ \hline
Majority baseline & .2145 & .3383 & .2096 (39.37) & .2127 & .3406 (.5166) & .3466 (.5305) & .3530 & .2427 (.3641) & .2234 & .4398 & .2921\\ 
Random baseline & .1907 & .4998 & .1815 (32.09) & .3374 & .4864 (.4923) & .5011 (.5052) &  .4830 & .3061 (.3769) & .1804 & .4652 & .3632\\ 
\model{State-of-the-art} & .5730$^a$ & - & .6110$^b$ (86.66)$^c$ & - & (.5470)$^d$ & .7995$^e$ & - & (.7104)$^f$ & \textbf{.6187}$^g$ & - & -\\
\model{BERT$_{SDL}$} & .6480 & .6167 & .7466 (88.57) & .3167 & .5347 (.5429) & .8012 (.8026) & .5699 & .6839 (.7018) & .5364 & .7274 & .6181 \\
\model{MT-DNN$_{SDL}$} & .6324 & .6019 & \textbf{.7690 (88.82)} & .3329 & .7066 (.7116) & \textbf{.8480 (.8486)} & .6211 & .6882 (.7080) & .5649 & .7506 & .6516 \\
\model{BERT$_{MDL}$} & \textbf{.6583} & .6157 & .7475 (88.60) & .3781 & .7211 (.7240) & .8093 (.8102) & .6444 & .6979 (.7162) & .5712 & .7414 & .6585\\
\model{MT-DNN$_{MDL}$} & .6526 & \textbf{.6174} & .7522 (88.85) & \textbf{.3797} & \textbf{.7772 (.7787)} & .8374 (.8383) & \textbf{.6541} & \textbf{.6979 (.7181)} & .5732 & \textbf{.7532} & \textbf{.6695}\\ \hline
Human Performance & .7730 & - & .7540 & - & - & (0.9090) & - & - & - & - & -\\ \Xhline{2\arrayrulewidth}
\end{tabular}
}
\caption{Results of experiments on all datasets in F$_1$m$_+$ (F$_1$ macro) and original paper metrics in parentheses (F$_1$m$_-$ (F$_1$ micro), Accuracy (Acc), Fake News Challenge score (FNC1), F$_1$ macro without class \textit{none} (F$_1$m$_+\setminus$ none)). $^a$TalosComb \cite{hanselowski2018fnc}; $^b$ESIM w/ GRU + Dropout \cite{Jiang2019fnc}; $^c$Ranking-MLP \cite{Zhang:2018}; $^d$Unigrams SVM \cite{barhaim2017stance}; $^e$BERT$_{CONS}$ \cite{popat-etal-2019-stancy}; $^f$TGMN-CR \cite{wei2018target}; $^g$GPT-based \cite{yang-etal-2019-blcu}.}
\label{table_stance_results}
\end{table*}

\section{Analysis}\label{sec:analysis}
As the robustness of an ML model is crucial if applied to other domains or in downstream applications, we analyze this feature in more detail. First, we define adversarial attacks to probe for weaknesses in the models. Second, we investigate the reason for detected weaknesses and a surprising anomaly in robustness between SDL and MDL models.

\subsection{Adversarial Attacks: Definition}
We investigate how robust the trained models are and whether TL from related tasks and MDL influence this property. Inspired by stress tests for NLI, we select three adversarial attacks to probe the robustness of the models and modify all samples of all test sets with the following configurations:

\noindent \textbf{Paraphrase} We paraphrase all samples of the test sets. For this, we lean on the work of \citeauthor{mallinson2017paraphrasing} \shortcite{mallinson2017paraphrasing} and train two machine translation models with OpenNMT \cite{opennmt}: one that translates English originals to German and another one that backtranslates.

\noindent  \textbf{Spelling} Spelling errors are quite common, especially in data from social media or debating forums. We add two errors into each input of a sample \cite{naik-etal-2018-stress}: (1) we swap two letters of a random word and (2) for a different word, we substitute a letter for another letter close to it on the keyboard. We only consider words with at least four letters, as shorter ones are mostly stopwords.

\noindent  \textbf{Negation} We use the negation stress test proposed by \citeauthor{naik-etal-2018-stress} \shortcite{naik-etal-2018-stress}. They add the tautology ``and false is not true'' after each sentence, as they suspect that models might be confused by strong negation words like ``not''. 
We assume the same is also valid for \ac{sd}. We add the tautology at the beginning of each sentence, since we truncate all inputs to a maximum length of 100 sub-words.



To measure the effectiveness of each adversarial attack $a \in A$, we calculate the potency score introduced by \cite{thorne-etal-2019-evaluating} as the average reduction from a perfect score and across the systems $s \in S$:

\begin{equation*}
    {\scriptstyle Potency(a) = c_a \dfrac{1}{|S|} \sum_{s \in S} (1-f(s, a))},
\end{equation*}

\begin{table}[!b]
\centering
\resizebox{0.40\textwidth}{!}{
\def\arraystretch{1.3}
\begin{tabular}{lccc}
\Xhline{2\arrayrulewidth}
\textbf{Method} & \textbf{\makecell[c]{Raw potency\\(\%)}} & \textbf{\makecell[c]{Correctness\\ratio}} & \textbf{\makecell[c]{Potency\\(\%)}}\\\hline
\textbf{\model{Spelling}} & 43.3 & 0.584 & 25.3\\
\textbf{\model{Negation}} & 41.1 & 1.0 & 41.1\\
\textbf{\model{Paraphrase}} & 38.0 & 0.632 & 24.0\\
\Xhline{2\arrayrulewidth}
\end{tabular}
}
\caption{Potency of all adversarial attacks.}
\label{table_stance_adversarial_summary_potency}
\end{table}

\noindent with c$_a$ being the ratio of correctly transformed samples (test to adversarial) and a function $f$ that returns the performance score for a system $s$ on an adversarial attack set $a$.

The correct rate c$_a$ is calculated by taking 25 randomly selected samples from all test sets and comparing them to their adversarial counterpart. For the paraphrase attack, the first author checked whether the paraphrased and original sentences are semantically equal.
We find that in 63\% of the samples this is the case.
This low result is mostly due to the three outlier datasets fnc1 (36\%), snopes (36\%), and arc (44\%).
Leaving out these three, 82\% of the sentences are semantically correct paraphrases.
As the changes through the spelling attack are minor and subjective to evaluate, we use the Flesch–Kincaid grade level \cite{kincaid1975derivation} to compare the readability of the original and adversarial sentences and label a sample as incorrectly translated if the readability of the adversarial sentence requires a higher U.S. grade level. For the negation attack samples, we assume a correctness of 100\% ($c_a = 1.0$) as the perturbation adds a tautology and the semantics and grammar are preserved.

\begin{table}[!t]
\centering
\resizebox{0.37\textwidth}{!}{
\def\arraystretch{1.3}
\begin{tabular}{lll}
\Xhline{2\arrayrulewidth}
 & \textbf{\model{BERT$_{SDL}$}}  & \textbf{\model{MT-DNN$_{MDL}$}}  \\ \hline
\textbf{Test} & \makecell[c]{.6181} & \makecell[c]{\textbf{.6695}}\\\hline
\textbf{\model{Spelling}} & .5568 \textbf{($-$9.9\%)} & \textbf{.5767} ($-$13.9\%)\\
\textbf{\model{Negation}} &\textbf{ .5914} \textbf{($-$4.3\%)} & .5871 ($-$12.3\%)\\
\textbf{\model{Paraphrase}} & .6012 \textbf{($-$2.8\%)} & \textbf{.6380} ($-$4.7\%)\\
\Xhline{2\arrayrulewidth}
\end{tabular}
}
\caption{Influence of adversarial attacks, averaged over all datasets on the \model{BERT$_{SDL}$} and \model{MT-DNN$_{MDL}$} model (in F$_1$m$_+$ and relative to the score on the test set).}
\label{table_stance_adversarial_summary}
\end{table}

\subsection{Adversarial Attacks: Results and Discussion}\label{sec:adv_attacks_discussion}

We choose to limit the compared systems to \model{BERT$_{SDL}$} and \model{MT-DNN$_{MDL}$}, as the latter uses both TL from related tasks and MDL, whereas the former uses neither. The potencies for all attack sets are shown in Table \ref{table_stance_adversarial_summary_potency} and ranked by the raw potency which assumes all adversarial samples to be correct (i.e. $c_a = 1$). The results on the adversarial attack sets for both the SDL and MDL model are shown in Table \ref{table_stance_adversarial_summary}.

The paraphrasing attack has the lowest raw potency of all adversarial sets and the average scores only drop by about 2.8-4.7\%.
Interestingly, on the datasets that turned out to be difficult to paraphrase (fnc1, arc, snopes), the score on the \model{MT-DNN$_{MDL}$} only drops by about 5.7\%, 6.4\%, and 6.5\% (see Appendix, Table \ref{table_stance_adversarial_abs}), which is not much below average.
This confirms \citet{niven-kao-2019-probing} in that the BERT architecture, despite contextualized word embeddings, also primarily focuses on certain cue words and the semantics of the whole sentence is not the main criterion.

With raw potencies of 41.1\% and 43.3\%, the negation and spelling attacks have the highest negative influence on both SDL and MDL (4.3\% to 13.9\% performance loss). We assume this to be another indicator that the models rely on certain key words and fail if the statistical occurrence of these words in the seen samples is changed. This is easy to see for the negation attack, as it adds a strong negation word.
For the spelling attack, we look at the following original example from the perspectrum dataset:

\begin{quote}
\textbf{Claim}: \textit{School Day Should Be Extended} \\
\textbf{Perspective}:  \textit{So much easier for parents!}\\
\textbf{Predict/Gold}:  \textit{support/support}
\end{quote}

\noindent And the same example as spelling attack:

\begin{quote}
\textbf{Claim}: \textit{School Day Sohuld Be Ectended} \\
\textbf{Perspective}:  \textit{So much esaier for oarents!}\\
\textbf{Predict/Gold}:  \textit{undermine/support}
\end{quote}

Since all words of the original sample are in the vocabulary, Google's sub-word implementation WordPiece \cite{wu2016google} does not split the tokens into sub-words. However, this is different for the perturbed sentence, as, for instance, the tokens ``esaier'' and ``oarents'' are not in the vocabulary. Hence, we get {[}esa, \#\#ier{]} and {[}o, \#\#are, \#\#nts{]}. These pieces do not carry the same meaning as before the perturbation and the model has not learned to handle them.

However, the most surprising observation represents the much higher relative drop in scores between the test and adversarial attack sets for MT-DNN$_{MDL}$ as compared to BERT$_{SDL}$. MDL should produce more robust models and support them in handling at least some of these attacks, as some of the datasets originate from Social Media and debating forums, where typos and other errors are quite common. On top of that, the model sees much more samples and should be more robust to paraphrased sentences.
Hence, to further evaluate the robustness of the two systems, we leverage the resilience measure introduced by \citet{thorne-etal-2019-evaluating}:

\begin{equation*}
    {\scriptstyle Resilience(s) =  \dfrac{\sum_{a \in A}c_a \times f(s, a)}{\sum_{a \in A} c_a}} .
\end{equation*}

It defines the robustness of a model against all adversarial attacks, scaled by the correctness of the attack sets. Surprisingly, the resilience of both the MDL (59.9\%) and SDL (58.5\%) model are almost on par.
The score, however, only considers the absolute performance on the adversarial sets, but not the drop in performance when compared to the test set results.
If, for instance, model \textit{A} performs better than model \textit{B} on the same test set, but has a higher drop in performance on the same adversarial set, model \textit{A} should show a lower robustness and thus receive a lower resilience score.
As the resilience score does not consider this, we adapt the equation by taking the performance of the test set $t$ into account:
\begin{equation*}
   {\scriptstyle Resilience_{rel}(s) = \dfrac{\sum_{a \in A}c_a \times (1 - f(s, t) - f(s, a))}{\sum_{a \in A} c_a}}.
\end{equation*}

\begin{table}
\centering
\resizebox{0.32\textwidth}{!}{
\def\arraystretch{1.3}
\begin{tabular}{lccc}
\Xhline{2\arrayrulewidth}
\textbf{Method} & \textbf{\makecell[c]{BERT$_{SDL}$}} & \textbf{\makecell[c]{MT-DNN$_{MDL}$}}\\\hline
\textbf{\model{Spelling}} & \textbf{98.4\%} & 97.6\% \\
\textbf{\model{Negation}} & \textbf{98.8\%} & 96.3\% \\
\textbf{\model{Paraphrase}} & \textbf{99.5\%} & 99.1\%\\\hline
\textbf{\model{Overall}} & \textbf{96.7\%} & 92.9\%\\
\Xhline{2\arrayrulewidth}
\end{tabular}
}
\caption{Resilience$_{rel}$ of BERT$_{SDL}$ and MT-DNN$_{MDL}$.}
\label{table_res_abs}
\end{table}

We calculate the score for all adversarial attacks separately, as well as the overall Resilience$_{rel}$, and observe that the SDL model outperforms the MDL model in each case (see Table \ref{table_res_abs}). For some datasets, the absolute F$_1$m$_+$ of the MDL model even drops below that of the SDL model (see Appendix, Table \ref{table_stance_adversarial_abs}). Our experiments show that performance-wise, we can benefit from MDL, but there is a high risk of drastic loss in robustness, which can cancel out the performance gains or, even worse, renders the model inferior in real-world scenarios. 
\subsection{Analysis of Robustness via Low Resource Experiments}
To investigate the reasons why the MDL model shows a lower robustness than the SDL models on average, we conduct low resource experiments by training the MDL model and the SDL models on 10, 30, and 70\% of the available training data.
Dev and test sets are kept at 100\% of the available data at all times and results are averaged over five seeds.


\begin{table}[!b]
    \centering
    \resizebox{0.45\textwidth}{!}{
\begin{tabular}{lcccc}
\Xhline{2\arrayrulewidth}
\textbf{Model / Ratio}  & \textbf{\makecell[c]{10\%}} & \textbf{\makecell[c]{30\%}} & \textbf{\makecell[c]{70\%}} & \textbf{\makecell[c]{100\%}}\\\hline
\textbf{\model{MT-DNN$_{MDL}$}} & \textbf{.5855} & \textbf{.6317}& \textbf{.6624} & \textbf{.6695} \\
\textit{\makecell[r]{(diff.)}}& \textit{($-.0953$)} & \textit{($-.0758$)} & \textit{($-.0598$)} & \textit{($-.0514$)}  \\
\textbf{\model{BERT$_{SDL}$}} & .4902 & .5559 & .6026 & .6181 \\
\Xhline{2\arrayrulewidth}
\end{tabular}
}
\caption{Train data ratio performance on the test set.}
\label{table_train_ratio_test}
\end{table}

\begin{figure}
\begin{subfigure}[c]{0.5\textwidth}
{\includegraphics[width=1.0\textwidth]
{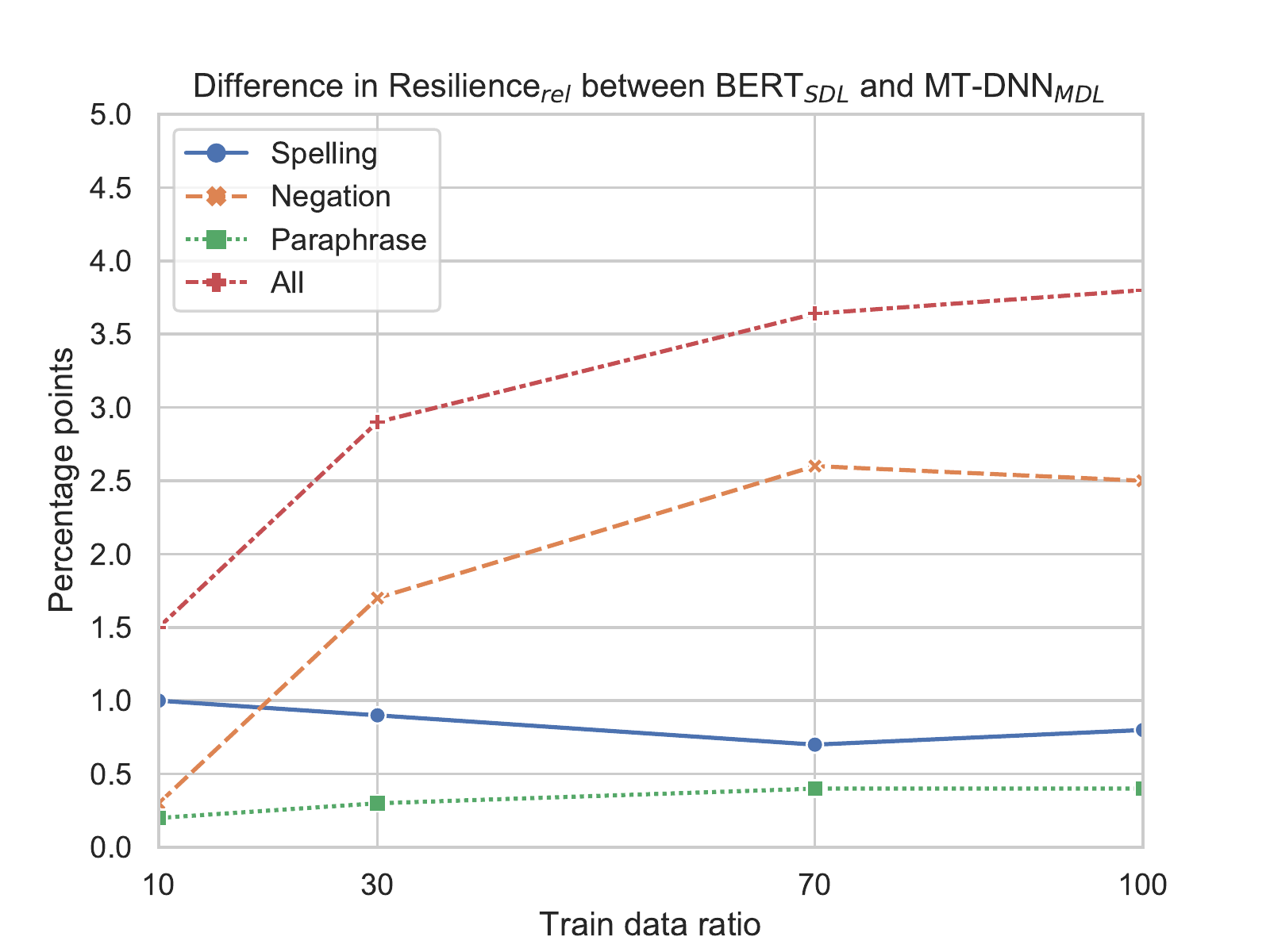}}

\subcaption{\label{table_train_ratio_res_fig} Difference in Resilience$_{rel}$ between BERT$_{SDL}$ and MT-DNN$_{MDL}$ for all train data ratios. }
\end{subfigure}
\begin{subtable}[c]{0.5\textwidth}
\centering
\resizebox{0.9\textwidth}{!}{
\def\arraystretch{1.3}
\begin{tabular}{lcccc}
\Xhline{2\arrayrulewidth}
\textbf{Models / Ratio}  & \textbf{\makecell[c]{10\%}} & \textbf{\makecell[c]{30\%}} & \textbf{\makecell[c]{70\%}} & \textbf{\makecell[c]{100\%}}\\\hline
\textbf{\model{MT-DNN$_{MDL}$}} & 97.0\% & 94.8\% & 93.4\% & 92.9\% \\
\textbf{\model{BERT$_{SDL}$}} & \textbf{98.5\%} & \textbf{97.7\%} & \textbf{97.0\%} & \textbf{96.7\%}  \\
\Xhline{2\arrayrulewidth}
\end{tabular}
}
\subcaption{\label{table_train_ratio_res}Overall Resilience$_{rel}$ for all train ratios.}
\end{subtable}
\caption{\label{fig_low_resource} Resilience$_{rel}$ over different train data ratios.}
\end{figure}

As is to be expected, the performance gap between \model{BERT$_{SDL}$} and \model{MT-DNN$_{MDL}$} on the test set grows with less training data (see Table \ref{table_train_ratio_test}). Here, the MDL shows its strength in low resource setups \cite{schulz18multi}.
Even more so, while the MDL model showed disencouraging performance w.r.t. adversarial attacks when trained on 100\% of the data, we observe that with less training data, the \model{MT-DNN$_{MDL}$ reduces the difference in overall Resilience$_{rel}$ to the \model{BERT$_{SDL}$} from 3.8pp at 100\% training data to 1.5pp at 10\% training data (see Table \ref{table_train_ratio_res}).
As shown in Figure \ref{table_train_ratio_res_fig}, this is due to the MT-DNN$_{MDL}$ approaching the Resilience$_{rel}$ of the BERT$_{SDL}$ against the negation and paraphrase attack.


Our analysis reveals that the amount of training data has a direct negative impact on model robustness. As most (if not all) datasets inevitably inherit the biases of their annotators \cite{geva-etal-2019-modeling}, we assume this negative impact on robustness is due to overfitting on biases in the training data. Hence, less training data leads to less overfitting on these biases, which in turn leads to a higher robustness towards certain attacks that target these biases.
For instance, the word ``not'' in the negation attack can be a bias that adheres to negative class labels \cite{niven-kao-2019-probing}.
Likewise, an overall shift in the distribution of some words due to the paraphrase attack can interfere with a learned bias.
We argue that spelling mistakes are unlikely to be learned as a bias for stance detection classes and the actual reason for the performance drop of the attack is due to the split of ungrammatical tokens into several sub-words (see section \ref{sec:adv_attacks_discussion}).

}
\section{Discussion and Future Work}\label{sec:discussion}
We introduced a StD benchmark system that combines TL and MDL and enables to add and evaluate adversarial attack sets and low resource experiments. We include ten \ac{sd} datasets of different domains into the benchmark and found the combination of TL and MDL to have a significant positive impact on performance. In five of the ten used datasets, we are able to show new state-of-the-art results. However, our analysis with three adversarial attacks reveals that, contrary to what is expected of TL and MDL, they result in a severe loss of robustness on our \ac{sd} datasets, with scores often dropping well below SDL performance. We investigate the reasons for this observation by conducting low resource experiments and conclude that one major issue is the overfitting on biases of vast amounts of training data in our MDL approach.

Reducing the amount of training data for both SDL and MDL models narrows down the robustness anomaly between these two setups, but also lowers the test set performance.
Hence, we recommend to develop methods that integrate de-biasing strategies into multi-task learning approaches---for instance, by letting the models learn which samples contain biases and should be penalized or ignored \cite{clark-etal-2019-dont} to enhance the robustness, thus also being able to leverage more (or all) training data available to maintain the performance. We foster this work by publishing our dataset splits, models, and experimental code.


In the future, we plan to combine methods that cope with biased data \cite{clark-etal-2019-dont,he-etal-2019-unlearn} with MDL and to experiment with sampling methods which aim to reduce the training data to the samples that are necessary to learn the task \cite{prabhu-etal-2019-sampling,ruder-plank-2017-learning}. 
In regard to adversarial attacks, we also aim to concentrate on task-specific adversarial attacks and use insights of adversarial attacks to build defences for the models \cite{pruthi-etal-2019-combating,wang2018defensive}.

\section*{Acknowledgments}
This work has been funded by the German Federal Ministry of Education and Research (BMBF) under the promotional reference 03VP02540 (ArgumenText).

\bibliography{acl2020}
\bibliographystyle{acl_natbib}

\appendix

\section{Appendices}
\label{sec:appendix}
\subsection{Adversarial Attacks on Stance Detection Models}
Table \ref{table_stance_adversarial_abs} shows the absolute performance scores of \model{MT-DNN$_{MDL}$} (all datasets with subscript $MDL$) and \model{BERT$_{SDL}$} (all datasets with subscript $SDL$). All absolute scores are in F$_1$ macro. The numbers in parentheses in the \textit{Avg.} column represent the relative drop to the respective score on the test set. Bold numbers in a column represent the best score between the $MDL$ and $SDL$ on an adversarial attack set.

\begin{table}[!htbp]
\centering
\resizebox{1.0\textwidth}{!}{
\def\arraystretch{1.3}
\begin{tabular}{lcccccccccc||c}
\Xhline{2\arrayrulewidth}
\textbf{Datasets} & \textbf{\makecell{arc\\12.4k}}  & \textbf{\makecell{argmin\\6.8k}} & \textbf{\makecell{fnc1\\42.5k}} & \textbf{\makecell{iac1\\4.2k}} & \textbf{\makecell{ibmcs\\0.9k}} & \textbf{\makecell{perspectrum\\7.0k}} &  \textbf{\makecell{scd\\3.3k}} & \textbf{\makecell{semeval2016t6\\2.5k}} & \textbf{\makecell{semeval2019t7\\5.2k}} &\textbf{\makecell{snopes\\14.4k}} & \textbf{Avg.} \\ \hline
\textbf{Test$_{SDL}$ }&.6480&.6167&.7466&.3167&.5347&.8012&.5699&.6839&.5364&.7274& .6182\\
\textbf{Test$_{MDL}$ }& \textbf{.6526}& \textbf{.6174}& \textbf{.7522}& \textbf{.3797}& \textbf{.7772}& \textbf{.8374}& \textbf{.6541}& \textbf{.6979}& \textbf{.5732}& \textbf{.7532}& \textbf{.6695}\\\hline

\model{\textbf{Negation$_{SDL}$}} & \textbf{.6463}&\textbf{.6205}&\textbf{.7233}&.3055&.5365&\textbf{.7854}&\textbf{.5962}&\textbf{.6799}&\textbf{.4266}&\textbf{.5942}&\textbf{.5914} \textbf{($-$4.3\%)}\\
\model{\textbf{Negation$_{MDL}$}} & .6398&.5832&.7017&\textbf{.3424}&\textbf{.6841}&.7497&.5901&.6550&.3358&.5896&.5871 ($-$12.3\%)\\\hline

\model{\textbf{Spelling$_{SDL}$}} & .4767&\textbf{.5863}&.6988&\textbf{.3492}&.4980&.6665&.5886&.5034&.5092&.6912&.5568 \textbf{($-$9.9\%)}\\
\model{\textbf{Spelling$_{MDL}$}} & \textbf{.4973}&.5403&\textbf{.7046}&.3311&\textbf{.6412}&\textbf{.6796}&\textbf{.6348}&\textbf{.5049}&\textbf{.5197}&\textbf{.7132}&\textbf{.5767} ($-$13.9\%)\\\hline


\model{\textbf{Paraphrase$_{SDL}$}}& .6043& \textbf{.6097}& .7019& .3477& .5320& .7649& .5735& .6589& .5137& \textbf{.7049}& .6012 \textbf{($-$2.8\%)}\\
\model{\textbf{Paraphrase$_{MDL}$}} & \textbf{.6110}& .6031& \textbf{.7093}& \textbf{.3682}& \textbf{.7489}& \textbf{.7941}& \textbf{.6321}& \textbf{.6611}& \textbf{.5483}& .7040& \textbf{.6380} ($-$4.7\%)\\
\Xhline{2\arrayrulewidth}
\end{tabular}
}
\caption{Comparison of \model{MT-DNN$_{MDL}$} (all datasets with subscript $MDL$) and \model{BERT$_{SDL}$} (all datasets with subscript $SDL$). All absolute scores are in F$_1$ macro.}
\label{table_stance_adversarial_abs}
\end{table}

\end{document}